\documentclass[sigconf,natbib=false,anonymous=false]{acmart}

\usepackage{multirow}
\usepackage{float}
\usepackage{placeins}
\usepackage{algorithm}
\usepackage{algpseudocode}
\usepackage{bbold}
\usepackage{libertine}
\usepackage{hyperref}

\AtBeginDocument{%
  }

\renewcommand\footnotetextcopyrightpermission[1]{} 

\RequirePackage[
  datamodel=acmdatamodel,
  style=acmnumeric,
  ]{biblatex}

\begin{document}

\title{Discrete Gene Crossover Accelerates Solution Discovery in Quality-Diversity Algorithms}

\author{Joshua L Hutchinson}
\affiliation{%
  \institution{University of Edinburgh}
  \city{Edinburgh}
  \country{United Kingdom}
}
\email{j.l.hutchinson@sms.ed.ac.uk}

\author{J.~Michael Herrmann}
\affiliation{%
  \institution{University of Edinburgh}
  \city{Edinburgh}
  \country{United Kingdom}
}
\email{michael.herrmann@ed.ac.uk}

\author{Sim\'on C. Smith}
\affiliation{%
  \institution{Edinburgh Napier University}
  \city{Edinburgh}
  \country{United Kingdom}}
\email{s.smith2@napier.ac.uk}

\renewcommand{\shortauthors}{Hutchinson et al.}

\begin{abstract}
Quality–Diversity (QD) algorithms aim to discover diverse, high-performing solutions across behavioral niches. However, QD search often stagnates as incremental variation operators struggle to propagate building blocks across large populations. Existing mutation operators rely on gradual variation to solutions, limiting their ability to efficiently explore regions of the search space distant from parent solutions or to spread beneficial genetic material through the population. We propose a mutation operator which augments variation-based operators with discrete, gene-level crossover, enabling rapid recombination of elite genetic material. This crossover mechanism mirrors the biological principle of meiosis and facilitates both the direct transfer of genetic material and the exploration of novel genotype configurations beyond the existing elite hypervolume. We evaluate operators on three locomotion environments, demonstrating improvements in QD score, coverage, and max fitness, with particularly strong performance in later stages of optimization once building blocks have been established in the archive. These results show that the addition of a discrete crossover mutation provides a complementary exploration mechanism that sustains quality-diversity growth beyond the performance demonstrated by existing operators. 
\end{abstract}

\ccsdesc[500]{Computing methodologies~Genetic algorithms}

\keywords{Quality-Diversity, MAP-Elites, Evolution}

\maketitle

\pagestyle{plain}

\section{Introduction}

Evolution by natural selection has given rise to a truly remarkable diversity of life on earth, with organisms occupying all known ecological niches~\cite{hutchinson_concluding_1957}. Quality-Diversity (QD) algorithms mirror natural evolution's ability to produce both specialization and diversity by maintaining populations of elite solutions across behaviorally distinct niches~\cite{mouret2015illuminating, pugh2016quality}. Among QD approaches, MAP-Elites has emerged as a foundational algorithm, due to its simplicity of implementation and alignment to QD principles \cite{mouret2015illuminating}. MAP-Elites discretizes the target behavior space into niches, maintaining only the highest fitness solution within each niche. The optimization process produces a population archive that illuminates the fitness landscape across behavioral dimensions.

The success of MAP-Elites, and related QD algorithms~\cite{pugh2016quality}, relies on the choice of variation operators. These operators determine the mechanisms through which parent genotypes are mutated or recombined to generate offspring. To the best of our knowledge, QD search has generally relied on the combination of two variation operators. First, isotropic Gaussian mutation~\cite{mouret2015illuminating}, which adds undirected random noise to parent solutions. And second, recombination by distance dependent variation between parent solutions~\cite{vassiliades2018discovering}. 

However, these operators have important limitations. When beneficial features emerge, these can only be propagated through the population via multiple sequential rounds of recombination. This gradual transfer limits the rate of transmission, thereby restricting exploration. At a population level, discovering novel combinations of features that already exist separately in the archive, but have never been jointly expressed, requires fortuitous sequences of intermediate mutations.

Our main hypothesis is that discrete crossover of parent genotypes can address these limitations through two complementary mechanisms. First, crossover enables rapid transport of successful genetic material across the population by directly exchanging genes between parents~\cite{wrro93026, back1993overview}, bypassing the restriction of sequential variation. Second, it promotes the exploration of genotypes that lie outside the existing elite hypervolume by frequently creating novel combinations of features~\cite{UmbarkarSheth2015, wrro93026}. This approach mirrors sexually reproducing biological systems, where chromosomal recombination generates offspring that exhibit combinations of parental traits, accelerating both the propagation and novel arrangement of successful genetic features~\cite{Barton1998, Otto2002}.

This work introduces two variation operators that augment standard gradual variation operators with discrete crossover. We evaluate these operators on three continuous control tasks and compare performance against the baseline operators.

\section{Background}

\subsection{MAP-Elites}

Quality-Diversity (QD) optimization comprises a class of evolutionary algorithms that aim to discover collections of high-performing yet behaviorally diverse solutions, rather than a single global optimum~\cite{pugh2016quality}. Compared to traditional evolutionary computation methods that collapse search pressure toward one best-performing individual, QD methods balance performance maximization with diversity, thereby promoting exploration across a user-defined behavior space. This paradigm has proven effective in domains characterized by multi-modality, deceptive fitness landscapes, or domains requiring robustness and adaptability, such as robotics, design optimization, and reinforcement learning~\cite{cully2015robots,ecoffet2021first,fontaine2020quality,allard2023online}.

MAP-Elites (Multi-dimensional Archive of Phenotypic Elites) is a QD algorithm which biases the search for diversity through the concept of behavioral niches~\cite{mouret2015illuminating}. In MAP-Elites, behaviors of interest are defined by a set of behavior descriptors set by expert users. These descriptors characterize solution phenotypes for the axes of a grid-based archive that partitions the behavior space into distinct niches. Each niche stores at most one solution: the highest-performing individual observed for that region of behavior space, referred to as the elite.

Search in MAP-Elites proceeds by iteratively selecting elites from the archive as parents and generating offspring via mutation and recombination. Each offspring is evaluated both for fitness (quality) and for its behavior descriptors. The offspring is then assigned to the corresponding niche in the archive. If the niche is empty, the offspring is inserted. If the niche already contains an elite, replacement occurs only if the offspring exhibits higher fitness, otherwise it is discarded. Through an evolutionary loop, MAP-Elites fills the archive with increasingly high-quality solutions spanning a wide range of behaviors.

\begin{algorithm}[t]
\caption{{\sc CVT-MAP-Elites}}
\label{alg:cvt-map-elites}
\begin{algorithmic}
\Require Fitness function $f()$, variation operator $variation()$, initial population $X_{\text{initial}}$, number of centroids $k$, generations $G$

\State \textbf{Generate CVT centroids:} $\{c_1, \dots, c_k\}$ in behavior space
\State \textbf{Initialize archive:} $X \gets \emptyset$, $F \gets \emptyset$
\State \hspace{\algorithmicindent} $X \gets X_{\text{initial}}$, $F \gets f(X_{\text{initial}})$

\State \textbf{For generation = 1 to $G$:}
\State \hspace{\algorithmicindent} \textbf{Select parent(s) from archive:} $x_p \sim X$
\State \hspace{\algorithmicindent} \textbf{Generate offspring:} $x' \gets variation(x_p)$
\State \hspace{\algorithmicindent} \textbf{Evaluate fitness and behavior:} $f', b' \gets f(x')$ 
\State \hspace{\algorithmicindent} \textbf{Assign to niche:} $j \gets$ index of nearest centroid to $b'$
\State \hspace{\algorithmicindent} \textbf{If niche empty or better:} $F[j] = f'$, $X[j] = x'$

\State \Return $(X, F)$ \Comment{Return archive of elites}
\end{algorithmic}
\end{algorithm}

This approach does pose limitations. First, exploration is driven indirectly through uniform or heuristic parent selection from the archive, without explicit pressure toward underexplored or low-performing regions of the behavior space. As a result, early-discovered elites can dominate reproductive opportunities, leading to uneven coverage and slow improvement in sparsely populated or initially low-fitness niches. Second, the discretization of the behavior space imposes a fixed resolution that may poorly match the underlying structure of the domain, potentially wasting evaluations on irrelevant distinctions while undersampling meaningful variation. Finally, the replacement rule—strictly fitness-based within each niche—can prematurely discard stepping-stone solutions that are locally suboptimal yet evolutionarily valuable for discovering new behaviors.

Several extensions to MAP-Elites have been proposed to address exploration limitations, including adaptive resolution archives~\cite{fontaine2019mapping}, novelty-aware selection strategies~\cite{lehman2011evolving}, and hybridization with gradient-based learning~\cite{faldor2025synergizing}. Of particular relevance to this work is CVT-MAP-Elites~\cite{vassiliades2017using}, which replaces the uniform grid archive with a Centroidal Voronoi Tessellation (CVT) over the behavior space. By defining niches via Voronoi cells centered on a fixed set of centroids, CVT-MAP-Elites scales MAP-Elites to higher-dimensional behavior descriptors while maintaining a bounded archive size. In our work, we employ the CVT-MAP-Elites variant, shown in Algorithm \ref{alg:cvt-map-elites}.

\subsection{QD variation operators}

Variation operators, or genetic operators, typically take one of three forms and determine how offspring are produced from parents: 

\begin{itemize}
    \item \textit{Selection operators} bias the choice of parents, $\mathbf{x}_i^{(t)}$, towards solutions of higher fitness. MAP-Elites achieves this through elite preservation, such that parents are selected at random from the set of elite genotypes~\cite{BaeckFogelMichalewicz2000}. 
    \item \textit{Crossover operators} combine parent genetic material to produce offspring, aiming to retain characteristics of elite solutions. This can take the form of discrete or intermediate recombination~\cite{BaeckFogelMichalewicz2000, Back1996}.
    \item \textit{Mutation operators} are applied to parents by perturbing the genotype parameters in order to produce offspring. These perturbations are commonly achieved through the addition of Gaussian noise~\cite{BaeckFogelMichalewicz2000, Back1996}.
\end{itemize}

The original MAP-Elites work explores the parameter space by randomly selecting parent solutions and mutating them through the addition of isotropic Gaussian noise ({\sc Iso}) to each parameter in their genotype (Equation~\ref{eqn:gaussian-mutation})~\cite{mouret2015illuminating}. The Gaussian noise, $\mathcal{N}(0, I)$, is scaled in intensity using the parameter $\sigma_{\text{iso}}$.

\begin{equation}
    \mathbf{x}_i^{(t+1)} 
    = \mathbf{x}_i^{(t)} 
    + \sigma_{\text{iso}}\mathbf{\mathcal{N}(0, I)}
\label{eqn:gaussian-mutation}
\end{equation}

As noted in~\cite{vassiliades2018discovering}, behaviorally diverse elites often share large portions of their genotypes. Thus, they tend to be localized to a high fitness region of the genotype space known as the ``elite hypervolume''. Whilst effective for exploration, isotropic noise explores indiscriminately and ignores these useful similarities between different regions of the gene pool~\cite{vassiliades2018discovering}.

This observation was used in~\cite{vassiliades2018discovering} to introduce a distance dependent, intermediate crossover operator which biases offspring along the existing dimensions of the elite hypervolume ({\sc Line}{\sc DD}). In genotype dimensions with little variation, offspring are perturbed a small amount and remain near the hypervolume; whereas dimensions with larger variations are more aggressively explored through larger perturbations. The combination of {\sc Iso}+{\sc Line}{\sc DD} (Equation~\ref{eqn:line mutation}) has featured prominently in QD literature~\cite{vassiliades2018discovering, Christou2023IsoLineDD, FontaineNikolaidis2021}.

\begin{equation}
\textbf{x}_i^{(t+1)} = \textbf{x}_i^{(t)} + \sigma_{\text{iso}}\mathbf{\mathcal{N}(0, I)} + \sigma_{\text{line}} (\textbf{x}_j^{(t)} - \textbf{x}_i^{(t)})\mathcal{N}(0, 1)
\label{eqn:line mutation}
\end{equation}

Other methods include gradient-based transformation of the genotype~\cite{faldor2025synergizing}, where a reinforcement learning-like phase is added to the QD optimization loop. This learning is added as an extra layer over traditional operators, and successfully are able to improve overall fitness. In this work, we focus on traditional operators that can be applied to any QD algorithm without large modifications.

\subsection{Discrete crossover operators}

In contrast to existing variation operators, discrete crossover operators focus on the recombination of parent genetic material to produce offspring~\cite{BaeckFogelMichalewicz2000}. The overall aim of crossover is to retain beneficial characteristics from two or more parents by directly copying genes into an offspring. This approach mirrors the crossover mechanism that occurs during sexual reproduction in biology. 

To determine which genes to crossover, a mask is used to specify from which parent to select the gene. Common approaches to mask design focus on either individual genes (with a probability of selecting parent A or B) or the genotype as a whole, such as $k$-point crossover~\cite{BaeckFogelMichalewicz2000}. Treating the genotype as a whole aligns more closely to the idea of features / building blocks, thus is used in this paper.

\section{Methods}

\subsection{Discrete crossover operators}

Our hypothesis aligns with the elite hypervolume concept, but notes the limitations of existing operators in exploring it effectively. Namely, that elite genetic material requires multiple generations of successful propagation to combine into the population and that exploration does not fully make use of the genetic diversity of the gene pool, instead exploring within or near the hypervolume of existing solutions.

To address these limitations, we propose the use of a discrete crossover operator. Firstly, this addresses the limitation of gradual gene transfer through direct exchange between parents, conserving features and recombining in a single variation event. Further, at a population level, this mechanism operates analogously to a grid search over the existing elite gene pool, testing combinations of elite features without requiring intermediate mutations, thus addressing the second limitation.

\begin{algorithm}[t]
\caption{{\sc {\sc Iso}{\sc Line}{\sc Cross}}Variation}
\label{alg:isolinecross}
\begin{algorithmic}
\Require Parents $\mathbf{x}_i^{(t)}, \mathbf{x}_j^{(t)}$; parameters $\sigma_{\text{iso}}$, $\sigma_{\text{line}}$, $\lambda_{\text{cross}}$, $p_{\text{cross}}$

\State \textbf{Apply {\sc Iso}+{\sc Line}{\sc DD} mutation to both parents:}\hfill(Eq.~\ref{eqn:line mutation})
\State \hspace{\algorithmicindent} $\mathbf{x}_i^{(a)} \gets \textbf{x}_i^{(t)} + \sigma_{\text{iso}}\mathbf{\mathcal{N}(0, I)} + \sigma_{\text{line}} (\textbf{x}_j^{(t)} - \textbf{x}_i^{(t)})\mathcal{N}(0, 1)$
\State \hspace{\algorithmicindent} $\mathbf{x}_i^{(b)} \gets \textbf{x}_j^{(t)} + \sigma_{\text{iso}}\mathbf{\mathcal{N}(0, I)} - \sigma_{\text{line}} (\textbf{x}_j^{(t)} - \textbf{x}_i^{(t)})\mathcal{N}(0, 1)$

\State \textbf{Generate Crossover Mask:}
\State \hspace{\algorithmicindent} $\mathbf{m} \sim \text{Mask Generator}(\lambda_{\text{cross}})$  \hfill (Eq. \ref{masj_a}-\ref{mask_eq})

\State \textbf{Apply Crossover:}
\State \hspace{\algorithmicindent} \textbf{if} $u \sim \mathcal{U}(0,1) < p_{\text{cross}}$ 
\State \hspace{\algorithmicindent} \hspace{\algorithmicindent} $\mathbf{x}_i^{(t+1)} \gets \mathbf{m} \odot \mathbf{x}_i^{(a)} + (1 - \mathbf{m}) \odot \mathbf{x}_i^{(b)}$
\hfill(Eq.~\ref{eqn:meiosis + noise mutation})
\State \hspace{\algorithmicindent} \textbf{else} 
\State \hspace{\algorithmicindent} \hspace{\algorithmicindent} $\mathbf{x}_i^{(t+1)} \gets \mathbf{x}_i^{(a)}$

\State \Return $\mathbf{x}_i^{(t+1)}$
\end{algorithmic}
\end{algorithm}

We note that discrete crossover alone provides no means to explore beyond existing genetic material, requiring its combination with a mutation operator. We therefore introduce two hybrid operators that integrate discrete crossover with established approaches:

\begin{itemize}

    \item {\sc Iso}{\sc Cross} ({\sc Iso} + Crossover): Combines isotropic Gaussian mutation with discrete crossover. First, two parents are selected at random from the elite population, then mutation is applied using Gaussian noise (Equation \ref{eqn:gaussian-mutation}), followed by recombination using a mask. 
    
    \item {\sc Iso}{\sc Line}{\sc Cross} ({\sc Iso}+{\sc Line}{\sc DD} + Crossover): Combines cross-over with both isotropic Gaussian mutation and directional variation. First, two parents are selected at random from the elite population, then mutation is applied using both Gaussian noise and {\sc Line}{\sc DD} (Equation \ref{eqn:line mutation}), followed by recombination using a mask. 
\end{itemize}

\begin{figure}[t]
    \centering
    \includegraphics[width=1\linewidth]{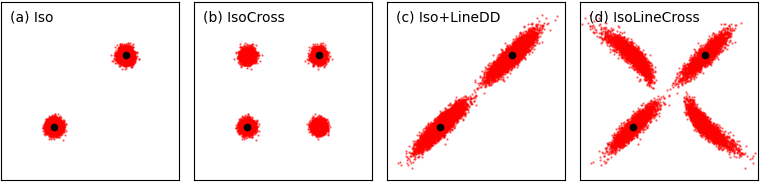}
    \caption{A 2D illustration of the baseline and proposed operators. Parents are shown in black with the order randomly selected for each sample. Offspring are shown in red. (A) Isotropic Gaussian mutation ({\sc Iso}). (B) {\sc {\sc Iso}{\sc Cross}}, demonstrating exploration through recombination. (C) {\sc Iso}+{\sc LineDD}, demonstrating directional variation between parents. (D) {\sc {\sc Iso}{\sc Line}{\sc Cross}}, demonstrating both directional variation between parents and exploration through recombination.}
    \Description{Parents are in the lower left and upper right of the plot. Panel A (Iso): Offspring are clustered around parents. Panel B (IsoCross): Offspring are clustered in region around parents and regions where parents are crossed over in the upper left and lower right. Panel C (Iso+LineDD): Offspring concentrated on the diagonal line between parents and lightly spread around each parent. Panel D (IsoLineCross): Offspring forming a combination of IsoCross and Iso+LineDD, such that there are two crossing diagonal lines, concentrated more densely on the original parent regions and the crossed over centers.}
    \label{fig:simple_simulation}
\end{figure}

The two operators implement the mutation procedure outlined in Algorithm \ref{alg:isolinecross} ({\sc Iso}{\sc Cross} sets $\sigma_{line} = 0$), first applying {\sc Iso}+{\sc Line}{\sc DD} using Equation~\ref{eqn:line mutation}, followed by probabilistic crossover 

\begin{equation}
\textbf{x}_i^{(t+1)} = \mathbf{m} \odot \textbf{x}_i^{(a)} + (1 - \mathbf{m}) \odot \textbf{x}_i^{(b)}
\label{eqn:meiosis + noise mutation}
\end{equation}

where $\textbf{x}_i^{(a)}$ and $\textbf{x}_i^{(b)}$ represent the perturbed parent solutions, $\textbf{m}$ the crossover mask, and $\odot$ the Hadamard element-wise product. Crossover is applied probabilistically to each solution, allowing control over the proportion of offspring generated through recombination, $p_{cross}$, versus {\sc Iso}+{\sc Line}{\sc DD} alone. Fig. \ref{fig:simple_simulation} illustrates the mutation patterns for the baseline and the proposed operators. Whilst both operators can produce two offspring per parent pair, we retain only the first offspring for archive insertion to maintain consistency with baseline operator implementations~\cite{chalumeau2023qdaxlibraryqualitydiversitypopulationbased}. 

\subsection{Crossover mask design}

Rather than using a predetermined crossover pattern, we generate masks stochastically using a Poisson process parametrized by crossover rate $\lambda_{cross}$~\cite{kingman_poisson_1992}. This design provides three key advantages: (i) biological plausibility, mirroring the approximately exponential spacing of crossover events during chromosome recombination~\cite{payero_crossover_2025, saito_regulation_2017}, (ii) scale invariance, preserving statistical properties across arbitrary genotype lengths~\cite{kingman_poisson_1992}, and (iii) problem-specific control over feature preservation (a smaller $\lambda$ will lead to fewer, more widely-spaced crossovers and longer preserved segments).

Given genotype length $N$, we first determine the number of crossover events (ensuring a minimum of one):
\begin{equation}
    K = \max(1, \lfloor \lambda \cdot N \rfloor)
    \label{masj_a}
\end{equation}

We then sample $K$ inter-crossover distances from an exponential distribution, setting the approximate intervals at which crossover will occur:
\begin{equation}
    d_i \sim \exp(1), \quad i \in \{1, \ldots, K\}
\end{equation}

Crossover positions are computed as $c_i = \sum_{j=1}^i d_j$ and linearly scaled to span the full genotype (enabling exploration throughout the genotype regardless of the exponential samples): 
\begin{equation}
    z_i = \left\lfloor \frac{c_i}{c_K} 
    (N-1) \right\rfloor.
\end{equation}

\begin{table*}[t]
\centering
\begin{tabular}{llcccc}
\toprule
Environment & Metric & {\sc Iso} & {\sc Iso}{\sc Cross} & {\sc Iso}+{\sc LineDD} & {\sc Iso}{\sc Line}{\sc Cross} \\
\midrule
\multirow{3}{*}{HalfCheetah Uni}
& QD Score ($\times 10^3$) &
$1{,}962 \pm 72$ &
$2{,}222 \pm 28$ &
$2{,}177 \pm 77$ &
$\textbf{2,273} \pm \textbf{29}$ \\
& Coverage (\%) &
$98.5 \pm 1.7$ &
$\textbf{100.0} \pm \textbf{0.0}$ & 
$\textbf{100.0} \pm \textbf{0.0}$ & 
$\textbf{100.0} \pm \textbf{0.0}$ \\
& Max Fitness ($\times 10^3$) &
$4.24 \pm 0.19$ &
$4.33 \pm 0.12$ &
$4.29 \pm 0.15$ &
$\textbf{4.36} \pm \textbf{0.14}$ \\

\midrule
\multirow{3}{*}{Hopper Uni}
& QD Score ($\times 10^3$) &
$825 \pm 97$ &
$849 \pm 60$ &
$841 \pm 73$ &
$\textbf{864} \pm \textbf{67}$ \\
& Coverage (\%) &
$79.0 \pm 1.2$ &
$79.5 \pm 0.5$ &
$\textbf{79.9} \pm \textbf{0.8}$ &
$\textbf{79.9} \pm \textbf{0.7}$ \\
& Max Fitness ($\times 10^3$) &
$1.34 \pm 0.09$ &
$1.44 \pm 0.15$ &
$1.39 \pm 0.15$ &
$\textbf{1.48} \pm \textbf{0.14}$ \\

\midrule
\multirow{3}{*}{Walker2d Uni}
& QD Score ($\times 10^3$) &
$660 \pm 72$ &
$681 \pm 52$ &
$820 \pm 64$ &
$\textbf{908} \pm \textbf{70}$ \\
& Coverage (\%) &
$65.7 \pm 6.3$ &
$72.5 \pm 2.6$ &
$78.8 \pm 2.5$ &
$\textbf{80.9} \pm \textbf{1.8}$ \\
& Max Fitness ($\times 10^3$) &
$1.46 \pm 0.16$ &
$1.40 \pm 0.12$ &
$1.59 \pm 0.14$ &
$\textbf{1.74} \pm \textbf{0.25}$ \\

\bottomrule
\end{tabular}
\vspace{0.4cm}

\caption{Mean $\pm$ St.d for QD Score, Coverage, and Max Fitness for {\sc Iso}, {\sc Iso}{\sc Cross}, {\sc Iso}+{\sc LineDD}, and {\sc Iso}{\sc Line}{\sc Cross} for each environment. Each experiment is replicated 20 times with random seeds. \textbf{Bold} indicates the highest performing operator(s).}
\label{tab:mean+std}
\end{table*}

Finally, the binary mask $\mathbf{m} \in \{0,1\}^N$ is constructed by counting the number of crossover points preceding each gene position and alternating accordingly:

\begin{equation}
    m_j = 1-\left[\left(\sum_{i=1}^K \mathbb{1}[j \geq z_i]\right) \bmod 2\right], \,\,\, j\in\{1,\ldots,N\}.
    \label{mask_eq}
\end{equation}

\subsection{Task environments}
\label{sec: envs+metrics}

We evaluate all operators on three continuous control environments from the Brax locomotion suite~\cite{freeman2021brax}: HalfCheetah Uni, Hopper Uni, and Walker2d Uni. We use identical hyperparameters across all environments (Appendix~Table~\ref{tab:hyperparameters}) and evaluate each operator configuration over 20 independent runs with different random seeds. Experiment parameters are listed in Appendix Table \ref{tab:experimental_setup}.

We score operators using standard QD metrics: QD score, coverage, and max fitness~\cite{FlageatLim2022}. QD score sums the fitnesses of all solutions in the archive (normalizing so all are $\geq 0$), providing an indicator of both quality (high fitness) and diversity (number of niches filled). Coverage calculates the percent of niches populated, capturing diversity. Max fitness represents the highest fitness achieved by any elite in the archive.

\section{Results}
\label{sec: results}

\subsection{Performance comparison}
\label{sec:performance comparison}

\begin{figure*}[t]
    \centering
    \includegraphics[width=1\linewidth]{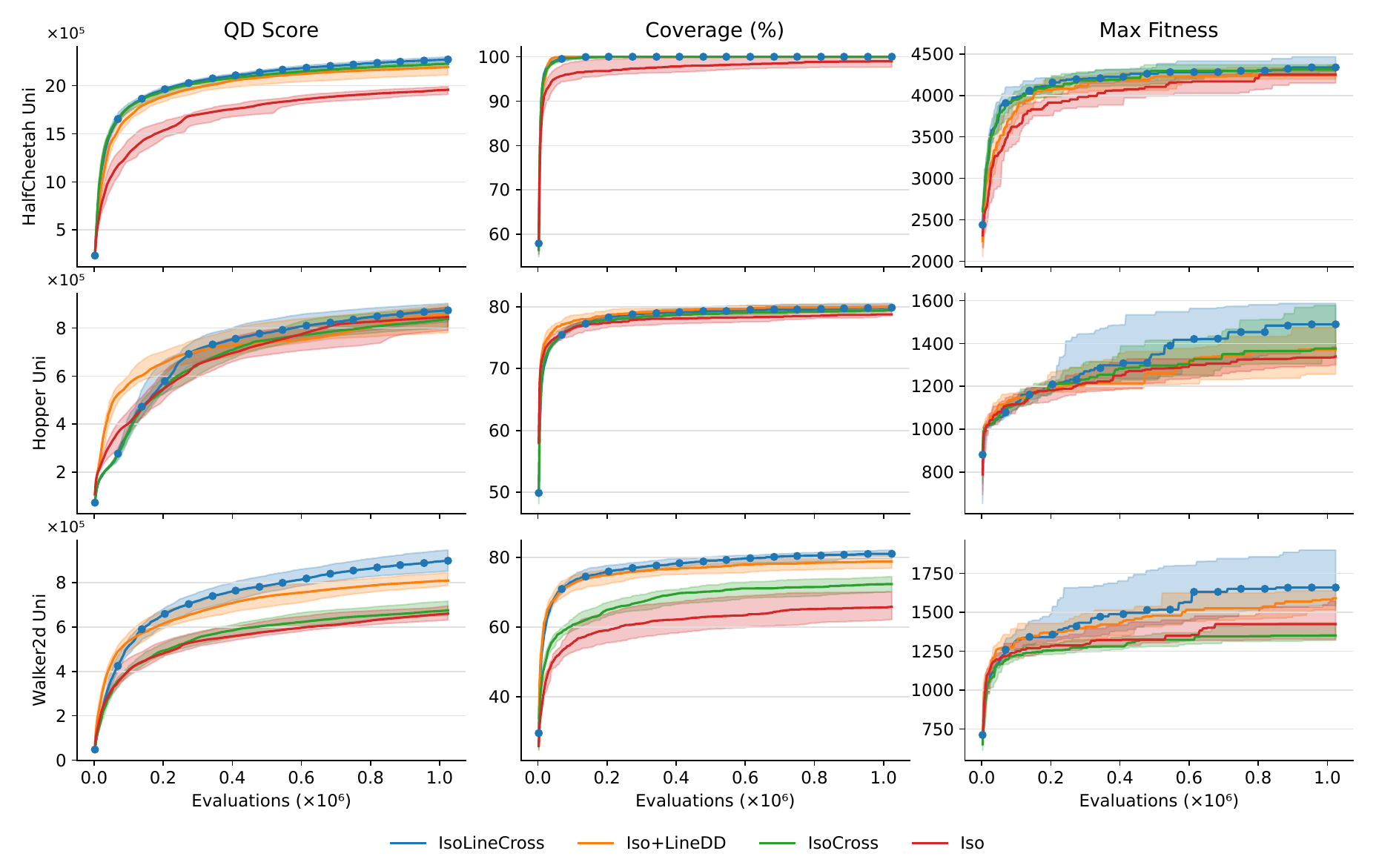}
    \caption{QD Score, Coverage, and Max Fitness for {\sc Iso}, {\sc Iso}{\sc Cross}, {\sc Iso}+{\sc LineDD}, and {\sc Iso}{\sc Line}{\sc Cross}. Each experiment is replicated 20 times with random seeds. The solid line is the median and the shaded region represents the interquartile range.}
    \Description{QD metrics for each operator and environment.}
    \label{fig:qd_metrics}
\end{figure*}

For each task, we evaluate the operators against the standard QD metrics: QD score, coverage, and max fitness (see Section \ref{sec: envs+metrics}). We report the mean and standard deviations in Table~\ref{tab:mean+std} and plot the median and interquartile range in Fig.~\ref{fig:qd_metrics}. Supplementary Fig.~\ref{fig:reproducibility} plots the distribution of results at the end of training. Sample archives showing the median QD score archive for each operator are shown in Supplementary Fig. \ref{a} and Fig. \ref{b}.

On all metrics and environments, we see that {\sc Iso}{\sc Line}{\sc Cross} achieves the highest or joint highest performance. Specifically, we see a significant increase in QD score achieved by {\sc Iso}{\sc Line}{\sc Cross} over {\sc Iso}+{\sc Line}{\sc DD} on both HalfCheetah and Walker2d ($p < 0.001$ on a Wilcoxon signed-rank test~\cite{wilcoxon1945}, with mean increases of $4.6\%$ and $11.3\%$ respectively). We see a similar improvement between {\sc Iso}{\sc Line}{\sc Cross} and {\sc Iso}{\sc Cross} on the same environments ($p < 0.001$, with mean increases of $2.3\%$ and $34.0\%$ respectively). On Coverage and Max Fitness, {\sc Iso}{\sc Line}{\sc Cross} also matches or exceeds {\sc Iso}+{\sc Line}{\sc DD} across all environments. Coverage increases by an average of $0.9\%$ ($p < 0.01$ on Walker2d) and Max Fitness by an average of $6.4\%$ ($p < 0.05$ on Hopper).

These findings show that the crossover mechanism does introduce useful variation during offspring production. {\sc Iso}{\sc Cross} shows near equivalent performance to {\sc Iso}+{\sc Line}{\sc DD} on both HalfCheetah and Hopper, demonstrating that crossover combined with {\sc Iso} is a valid exploration approach. However, it performs significantly worse on Walker2d, potentially indicating a frailty in certain environments. {\sc Iso}{\sc Line}{\sc Cross} shows no such deficiencies, indicating that it is the coupling of the locally operating, directional variation of {\sc Line}{\sc DD} with the globally operating discrete crossover that results in the increase in performance. 

\subsection{Offspring quality analysis}

\begin{figure}[t]
    \centering
    \includegraphics[width=1\linewidth]{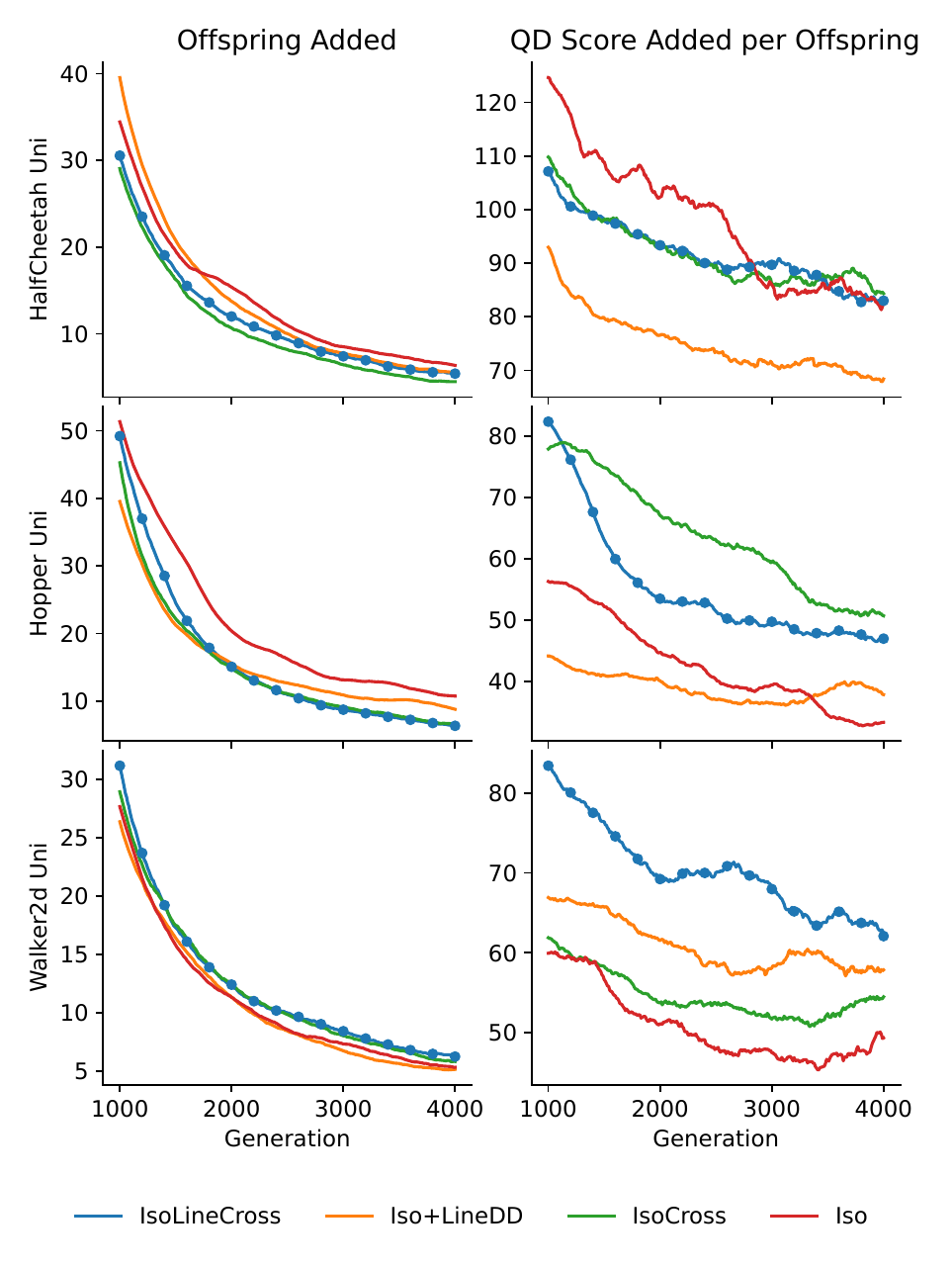}
    \caption{500 generation rolling average of offspring added (left) and the QD score added per offspring (right) for each operator and environment. We see similar levels of offspring added for all operators, with QD score added notably higher for the {\sc Iso}{\sc Line}{\sc Cross} operator than {\sc Iso}+{\sc Line}{\sc DD} across all generations.}
    \Description{Left: Decreasing rate of offspring addition over generations and approximately similar values for all operators. Right: Higher values of QD score added for {\sc Iso}{\sc Line}{\sc Cross} than other operators.}
    \label{fig:GA Metrics}
\end{figure}

In Walker2d, and, to a lesser degree, in Hopper, we observe a crossing of the {\sc Iso}+{\sc Line}{\sc DD} and {\sc Iso}{\sc Line}{\sc Cross} QD scores after the initial exploration phase. We explore this idea further in Fig. \ref{fig:GA Metrics}, where we compare the number of offspring and QD score per offspring added by each operator per generation.  

We observe that after the initial $\sim$1000 generations, the number of offspring added are fairly consistent between operators (Fig. \ref{fig:GA Metrics}, left). Thus, the main differentiator is the ability to produce higher quality offspring which contribute more to the increase in QD score. {\sc Iso}{\sc Line}{\sc Cross} again exhibits superior performance, consistently adding offspring of higher quality than other operators\footnote{We note that the {\sc Iso} operator on HalfCheetah is notably higher, however, this is due to the other operators having already reached 100\% coverage and thus saturated the ability to add QD score through behavioral novelty.}. Strikingly, {\sc Iso}{\sc Line}{\sc Cross} exceeds {\sc Iso}+{\sc Line}{\sc DD} in QD score added per offspring by an average of 21.1\% across all generations and environments. 

Combining these results with those of Fig. \ref{fig:qd_metrics} reveal characteristics of the operators. As evidenced by the rapid increase in QD score at the start of Walker2d and Hopper, {\sc Iso}+{\sc Line}{\sc DD} is particularly effective in early generations when the genotype space is sparsely explored. However, as the population matures, the discrete crossover mechanism becomes important in maintaining exploration and producing higher quality offspring.

We hypothesize that this shift in effectiveness is due to the absence of high quality genotype features in early generations, limiting the usefulness of exchange. However, once these have been established within the elite population, recombination enables higher quality exploration.

\subsection{Genotype space analysis}

To understand the mechanisms underlying the performance differences observed, we analyze the geometry of the final elite archives in genotype space. We quantify the effective dimensionality of each archive via principal component analysis, defining this as the number of principal components required to explain 95\% of the variance in the genotypes of the final elite population~\cite{JolliffeCadima2016PCA}.

Results across the three environments reveal a consistent pattern (Table \ref{tab:effective_rank}). The baseline operators, {\sc Iso} and {\sc Iso}+{\sc LineDD}, rely on a limited number of dimensions for variation, requiring between 4--11 principal dimensions, despite the $\sim$19,000-dimensional genotype space. In contrast, discrete crossover operators explore substantially higher-dimensional subspaces: {\sc Iso}{\sc Cross} requires 26--133 effective dimensions, while {\sc Iso}{\sc Line}{\sc Cross} requires 16--111 dimensions.

However, dimensional complexity does not directly correlate with performance. {\sc Iso}{\sc Cross}, despite requiring the highest-dimensional subspace in all environments, fails to consistently outperform the other operators, and notably performs closer to {\sc Iso} in Walker2d. This supports our hypothesis that discrete crossover is effective in promoting exploration, but does so inefficiently, with limited refinement within these dimensions.

{\sc Iso}{\sc Line}{\sc Cross} presents a middle ground. By combining the sustained  exploratory power of crossover with the intra-dimension directed variation from line mutation, it focuses exploration into a valuable, intermediate-dimensional subspace (4--11 times higher than {\sc Iso}+{\sc LineDD}, but lower than {\sc Iso}{\sc Cross}). Coupled with the greater quality of offspring shown in Fig. \ref{fig:GA Metrics}, this suggests that {\sc Iso}{\sc Line}{\sc Cross} explores to more dimensions and makes effective use of them to produce high quality offspring.

\begin{table}[t]
\centering

\begin{tabular}{llrr}
\toprule
Environment & Operator & Mean & St.d. \\
\midrule
\multirow{4}{*}{HalfCheetah} 

 & {\sc Iso}+{\sc LineDD} & 9.45 & 0.95 \\
  & {\sc Iso} & 11.25 & 1.33 \\

 & {\sc Iso}{\sc Line}{\sc Cross} & 111.65 & 10.51 \\
  & {\sc Iso}{\sc Cross} & 133.15 & 10.16 \\
\midrule
\multirow{4}{*}{Hopper} 

 & {\sc Iso}+{\sc LineDD} & 3.75 & 1.07 \\
  & {\sc Iso} & 6.75 & 1.97 \\

 & {\sc Iso}{\sc Line}{\sc Cross} & 16.15 & 9.59 \\
  & {\sc Iso}{\sc Cross} & 26.40 & 10.99 \\
\midrule
\multirow{4}{*}{Walker2d} 

 & {\sc Iso}+{\sc LineDD} & 4.65 & 1.27 \\
  & {\sc Iso} & 6.70 & 1.75 \\

 & {\sc Iso}{\sc Line}{\sc Cross} & 20.60 & 7.00 \\
  & {\sc Iso}{\sc Cross} & 40.80 & 74.45 \\
\bottomrule
\end{tabular}
\vspace{0.4cm}
\caption{Effective dimensionality measured by principal components required to explain 95\% of the variance in the genotypes of the final elite population. Values show mean $\pm$ st.d. over 20 random seeds.}
\label{tab:effective_rank}
\end{table}

\section{Discussion}

\subsection{Future refinements}

In this paper, we have expanded the set of QD variation operators and demonstrated an increase in performance over standard operators. Based on these results, we note several interesting areas for future refinement.

Our experiments maintained fixed hyperparameters across all environments with different operators exhibiting superiority at different evolutionary stages. The success of {\sc Iso}+{\sc LineDD} in early generations and {\sc Iso}{\sc Line}{\sc Cross} in later stages (Fig. \ref{fig:qd_metrics}) suggests that adaptive operator selection could improve performance. This raises the question that fine tuning the relative contribution of each operator at different evolutionary stages could further improve the optimization process. Such an approach was used by Gaier et al.~\cite{Gaier_2020}, using the multi-armed bandit algorithm UCB~\cite{sutton_reinforcement_2018} to select the mix of operators used for each generation. We propose to extend this to use both the probability of applying each operator and the intensity of variation. For instance, crossover probability could be increased as archive diversity grows and more building blocks become available for recombination. 

An alternative approach we would be interested to explore is the use of sporadic/periodic discrete crossover throughout the optimization process. Theoretically, this could enable phases of local refinement in high variance dimensions through directed variation, punctuated by phases of global exploration through the scattering of elite genetic material. Such an approach would mimic the evolutionary strategy of diverse plants, fungi, and microorganisms which alternate between sexual and asexual reproduction depending on environmental conditions~\cite{Barbuti2012Population, Dudgeon2017Switch}.

Another potential refinement would be the investigation of alternative crossover mask designs. Our approach used Poisson-distributed crossover points, mirroring chromosome recombination, and varied between offspring. However, it remains to be studied whether other recombination operators, such as fixed $k$-point crossover or crossover based on other vectors in the population (e.g., similar to Differential Evolution~\cite{price2005differential}), can improve upon the current results.

\subsection{Integration with other QD methods}

As demonstrated in Section \ref{sec: results}, {\sc Iso}{\sc Line}{\sc Cross} exhibits superior performance over existing operators on all environments. It can serve as a drop in replacement for existing operators\footnote{The operator is written as a very short extension to the QDax package~\cite{chalumeau2023qdaxlibraryqualitydiversitypopulationbased}}, requiring only two additional hyperparameters. Thus, we believe that it serves an important role in the QD toolkit and that any QD method using a genetic algorithm could potentially benefit from its use. 

As a result of the building block nature encouraged by discrete crossover, we believe it would be a particularly worthwhile extension to apply {\sc Iso}{\sc Line}{\sc Cross} to methods which use learned models in conjunction with the archive~\cite{faldor2025synergizing, Gaier_2020}. It may be the case that the building blocks benefit learnability and thus adaptability, with the modular blocks potentially offering an alternative route to behavior adaptation in changing environmental conditions.

\subsection{Evolvability and representation}

Our results also suggest a deeper phenomenon beyond immediate performance gains: the potential for archives to develop representations that are evolvable under the applied variation operators. The archive does not merely store elite solutions, it stores solutions that are \textit{elite under the specific variation operators used to generate them}. This creates implicit selection pressure favoring genotypes that produce valuable offspring when subjected to this type of variation - that they are both elite and \textit{evolvable}~\cite{WagnerAltenberg1996Evolvability}.

If this hypothesis holds, populations evolved with discrete crossover might exhibit an emergent modular structure, with parameter segments that function coherently when exchanged between solutions. Thus, solutions that align their functional modules with typical crossover boundaries would likely generate higher-quality offspring, gradually enriching the archive with increasingly evolvable genotypes. 

This connects to questions about representation in evolutionary algorithms. If populations co-adapt with their variation operators, then operator design becomes not just a search strategy but a mechanism shaping the evolved solutions themselves. The surprising effectiveness of discrete crossover on continuous neural network parameters, counter to initial intuition, may reflect precisely this phenomenon: the population learning to be crossover-compatible. 

More broadly, if mirroring of operator dynamics is true, then more open-ended evolutionary search processes might develop increasingly sophisticated internal representations that facilitate different forms of recombination. These questions point toward a research direction examining the co-evolution of genotypes and genetic operators. Understanding how populations adapt to their mutation mechanisms could inform both operator design and representation choices, ultimately enabling more effective evolutionary search across diverse problem domains.

\section{Conclusion} 

In this paper, we introduced two discrete crossover-based variation operators that can accelerate the recombination of successful building blocks among elite solutions. {\sc Iso}{\sc Cross} achieved performance equal to the current standard operator {\sc Iso}+{\sc Line}{\sc DD} in two of three environments, notably failing in the third.

{\sc Iso}{\sc Line}{\sc Cross}, the combination of discrete crossover with the existing operator, had no such failure and demonstrated a sustained performance improvement on all metrics and environments. The operator was particularly effective in maintaining exploration and the production of high quality offspring in later evolutionary stages.

\begin{acks}
Acknowledgments are suppressed whilst the paper is under review.
\end{acks}

\section*{Source code}
The source code for the operators and experiments can be found at \href{https://github.com/JoshuaLHutchinson/hutchinson_2026_gecco}{hutchinson\_2026\_gecco}

\printbibliography

\clearpage
\onecolumn

\appendix

\section{Supplementary results}

\subsection{Distribution of results}

\begin{figure*}[h]
    \centering
    \includegraphics[width=1\linewidth]{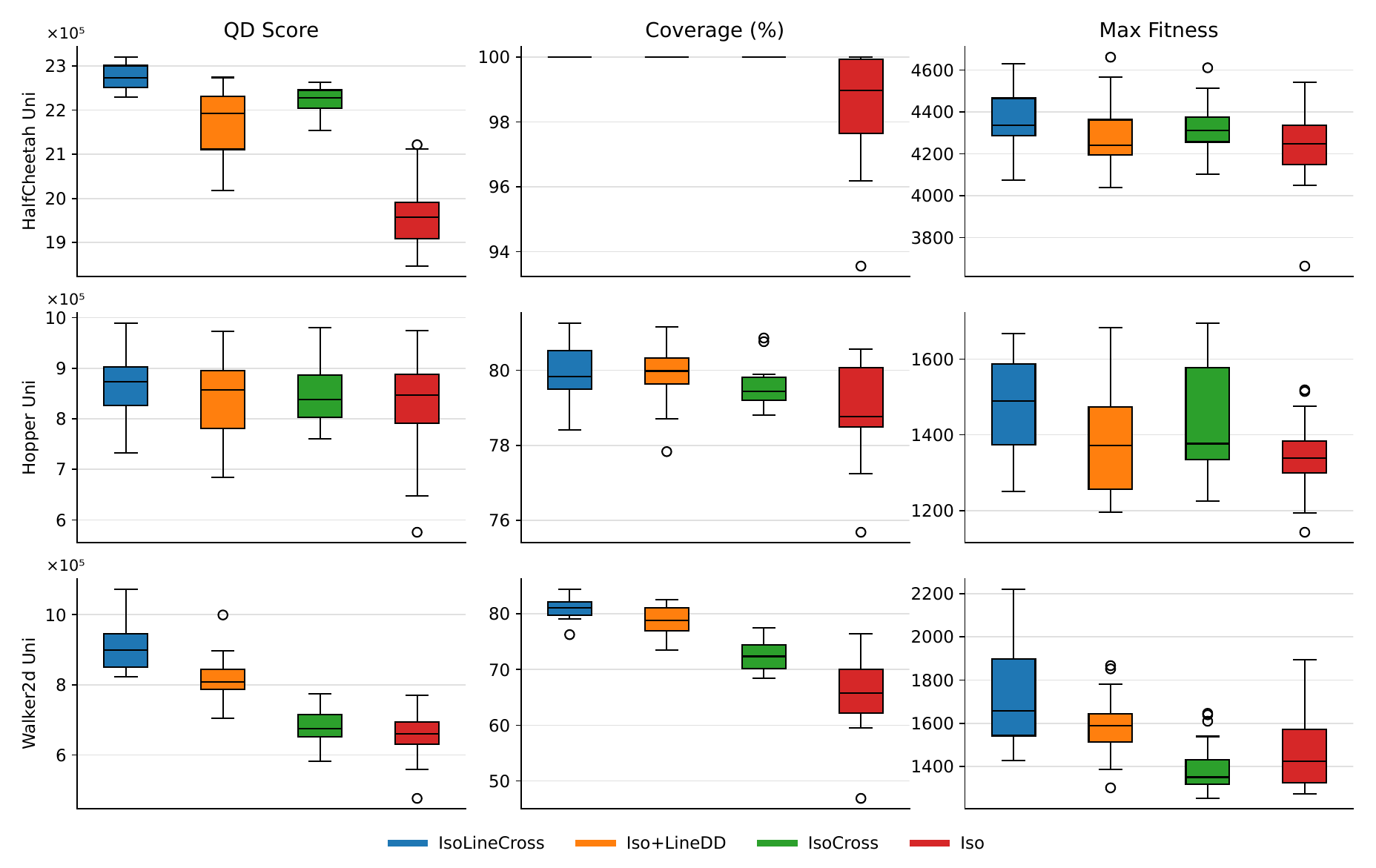}
    \caption{Box plots showing the distribution of results for each of the operators on the QD metrics (see Section \ref{sec: envs+metrics}). The black line within the box represents the median, the box the interquartile range, the whiskers 1.5$x$ IQR, and the white circle outliers beyond this range.}
    \Description{Box plots comparing the distribution of QD metrics achieved by each operator across 20 independent runs.}
    \label{fig:reproducibility}
\end{figure*}

\clearpage

\subsection{Archives}

Below we show sample archives for each operator and environment at the end of training. Each archive is chosen based on the seed with the QD score closest to the median QD score for the operator in that environment to present a representative sample.

\begin{figure}[h]
    \centering
    \includegraphics[width=1\linewidth]{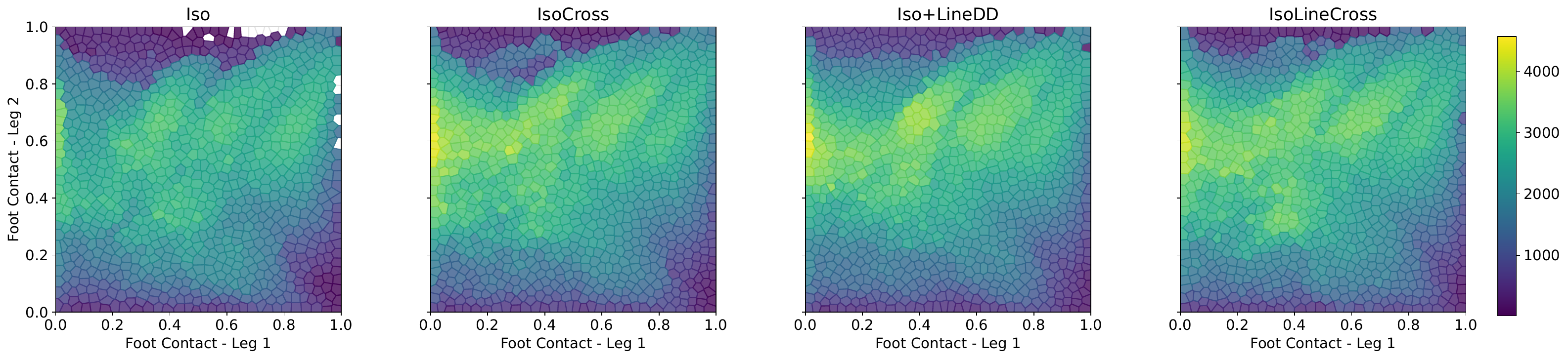}
    \caption{HalfCheetah Uni archive at the end of training for all operators.}
    \Description{MAP-Elites archives for HalfCheetah Uni showing approximately equivalent performance for all non-{\sc Iso} operators.}
    \label{a}
\end{figure}

\begin{figure}[h]
    \centering
    \includegraphics[width=1\linewidth]{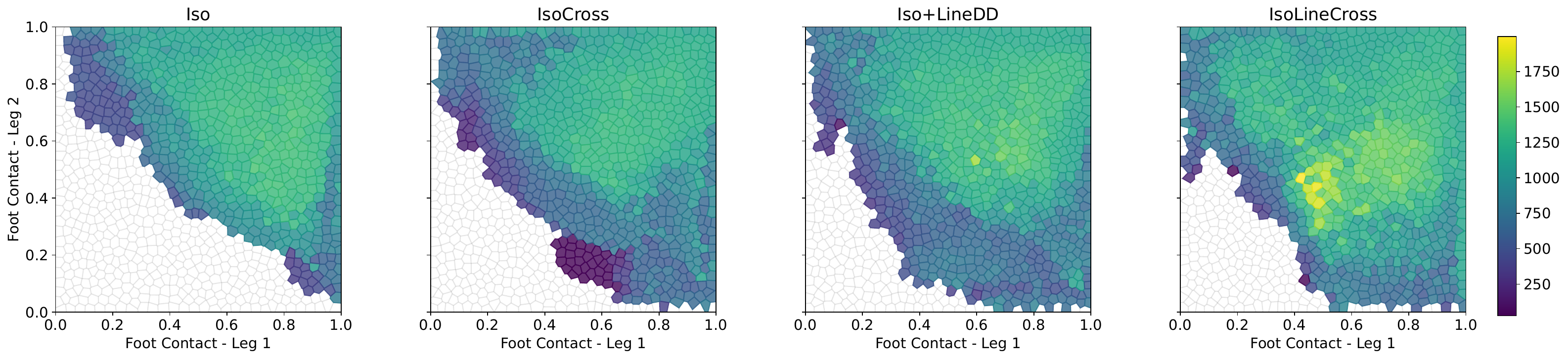}
    \caption{Walker2d Uni archive at the end of training for all operators.}
    \Description{MAP-Elites archives for Walker2d Uni showing superior performance of {\sc Iso}{\sc Line}{\sc Cross}.}
    \label{b}
\end{figure}

\clearpage

\section{Configuration}

\subsection{Variation operator hyperparameters}

\begin{table*}[h]
\centering
\begin{tabular}{lcccc}
\toprule
 & {\sc Iso} & {\sc {\sc Iso}{\sc Cross}} & {\sc Iso}+{\sc Line}{\sc DD}  & {\sc {\sc Iso}{\sc Line}{\sc Cross}} \\
\midrule
{\sc Iso} intensity $\sigma_{\text{iso}}$      & 0.005 & 0.005 & 0.005 & 0.005 \\
{\sc Iso} probability $p_{\text{iso}}$        & 1.0   & 1.0   & 1.0   & 1.0   \\
\midrule
{\sc Line}{\sc DD} intensity $\sigma_{\text{line}}$      & --    & --  & 0.05    & 0.05  \\
{\sc Line}{\sc DD} probability $p_{\text{line}}$      & --    & --   & 1.0    & 1.0   \\
\midrule
{\sc Cross} mask parameter $\lambda_{\text{cross}}$    & --    & 0.1    &  --  & 0.1   \\
{\sc Cross} probability $p_{\text{cross}}$       & --    & 0.5    & --   & 0.5   \\
\bottomrule
\end{tabular}
\vspace{0.4cm}
\caption{Variation operators with associated intensities and probabilities. These remain common across all experiments.}
\label{tab:hyperparameters}
\end{table*}

\subsection{Experiment setup}

\begin{table}[h]
\centering
\begin{tabular}{lccc}
\toprule
Parameter & HalfCheetah Uni & Hopper Uni & Walker2d Uni \\
\midrule
Environment name & halfcheetah\_uni & hopper\_uni & walker2d\_uni \\
Episode length & 1000 & 1000 & 1000 \\
Number of generations $G$ & 4000 & 4000 & 4000 \\
Batch size & 256 & 256 & 256 \\
Initial CVT samples & 50000 & 50000 & 50000 \\
Number of centroids & 1024 & 1024 & 1024 \\
Policy hidden layer sizes & [128, 128, $|\mathcal{A}|$] & [128, 128, $|\mathcal{A}|$] & [128, 128, $|\mathcal{A}|$] \\
\bottomrule
\end{tabular}
\vspace{0.4cm}
\caption{Parameters of the experimental setup used for QD experiments across HalfCheetah Uni, Hopper Uni, and Walker2d Uni environments.}
\label{tab:experimental_setup}
\end{table}

\clearpage

\end{document}